\documentclass[11pt]{article}

\usepackage[T1]{fontenc}
\usepackage[utf8]{inputenc}
\usepackage{libertinus}
\usepackage[libertinus]{newtxmath}
\usepackage[scaled=0.83]{beramono}
\usepackage{microtype}
\usepackage[a4paper,margin=1in]{geometry}
\usepackage{graphicx}
\usepackage{booktabs}
\usepackage{tabularx}
\usepackage{multirow}
\usepackage{array}
\usepackage[table]{xcolor}
\usepackage{etoolbox}
\usepackage{enumitem}
\usepackage[skins,breakable]{tcolorbox}
\usepackage[numbers,sort&compress]{natbib}
\usepackage{caption}
\DeclareCaptionType[fileext=lol]{listing}[Listing][List of Listings]
\usepackage{xurl}
\usepackage[hidelinks]{hyperref}
\usepackage[capitalise,noabbrev]{cleveref}
\crefname{listing}{Listing}{Listings}

\definecolor{vbblue}{HTML}{2a78d6}
\definecolor{pdorange}{HTML}{eb6834}
\definecolor{vccaqua}{HTML}{1baf7a}
\definecolor{ink}{HTML}{2b2a26}
\definecolor{muted}{HTML}{6f6d65}
\definecolor{panel}{HTML}{f4f3ee}
\definecolor{rowshade}{HTML}{f0efe9}

\hypersetup{colorlinks=true,linkcolor=ink,citecolor=vbblue,urlcolor=vbblue}
\AtBeginEnvironment{table}{\rowcolors{2}{}{rowshade}}  
\setlist{itemsep=2pt,topsep=4pt}
\newcommand{\vb}{\textsf{transparent-compact}}
\newcommand{\vbh}{\textsf{\shortstack{transparent-\\compact}}} 
\newcommand{\pd}{\textsf{pi-default}}
\newcommand{\vcc}{\textsf{pi-vcc}}
\newcommand{\bil}{\textsf{billion-context-pi}}
\newcommand{\qwen}{Qwen3.8-27B}
\newcommand{\luna}{GPT-6 Luna}
\newcommand{\sol}{GPT-6.1 Sol}
\newcommand{\gemma}{Gemma4 31B}

\title{\textbf{Maximising Transparency in Context Compaction Improves Retention of Facts and Commitments for Lightweight Coding Agents}}
\author{Felix Woolford\\[2pt]
\small Center for Human Nature, Artificial Intelligence, and Neuroscience (CHAIN)\\
\small Hokkaido University}
\date{}

\begin{document}
\maketitle

\begin{abstract}
\noindent
Long coding agent sessions outgrow the model's context window, so agents \emph{compact}: older history is replaced by something shorter, usually a model-written summary. A summary selects and rewords, and the agent cannot see what was lost.  
Existing work has explored a range of compaction techniques, but comparatively little attention has been paid to their effectiveness with lightweight models, including those deployed locally or used for high-volume workloads.
Here, we present \vb{}, a simple deterministic method for the open-source Pi coding agent, designed specifically for use with such models and with the purpose of maximising the transparency of the agent's view of its own compacted context.  
It guarantees that user and assistant prose in its context is unchanged by a model or algorithm, and explicitly flags where compaction has stripped information. We compare it with Pi's default model-written summary and with \vcc{}, a popular deterministic compaction method with similar methods but without verbatim retention of prose. 
Our benchmarks run on four models of varying performance levels, with emphasis on \qwen{}: a lightweight model at the top end of what consumer hardware can serve locally. On sets probing \emph{commitments} (rules the user set which exist only in the conversation) and recall of facts, \vb{} matched or outperformed both alternatives. 
In particular, it showed an advantage in avoiding confident blunders, i.e. assertion of false information as opposed to recognizing that something was not recalled.
Even with a frontier model, \vb{}'s advantage was still present but smaller, suggesting that method is broadly effective, although the strongest models are able to compensate for ambiguous summaries more effectively. We argue that \vb{} spares lightweight models from reasoning or searching their way around what compaction lost, and results in improved recall and application of critical information as a result, without the need for task or context-specific structuring of the session history.
\end{abstract}

\section{Introduction}
\label{sec:intro}

A coding agent works in long sessions. It reads files, runs commands and tests, and talks with the user about what to do. Most of what fills its context is tool output: about 84\% of an average SWE-agent turn \citep{lindenbauer2025complexity}, and input tokens, not output tokens, drive the cost of agentic work \citep{bai2026money}. When the context approaches the model's window, the agent \emph{compacts}. The usual approach is to ask the model to summarise the older history and continue from the summary \citep{anthropic2025context}. A summary is selective by design. It keeps what the summariser judged important, in the summariser's words, and nothing in the agent's new context marks what was dropped or changed.

Some losses can be repaired, for example the agent can re-read a file or re-run a test. But other losses are silently damaging: A user's \emph{commitments} (``don't touch the migration folder'', ``no need for backwards compatibility'', ``we agreed to keep it in one file'') may exist only in the conversation. If nothing in the repository reveals them, then an agent that has lost one can act against it without noticing. Recent work has shown that compaction silently erases such constraints: violations rose from 0\% with the policy in context to 30\% after compaction, and stayed at 0\% when the constraint survived the summary \citep{chen2026governance}.

Compaction methods are typically developed against the most capable frontier models, whose capabilities allow them to compensate for some of the limitations of lossy summaries.
This paper focuses instead on lighter weight models that are effective enough to have real-world utility, but whose weaker reasoning exposes them to failures deriving from compaction. A 27-billion-parameter model such as \qwen{} now runs on a single consumer GPU and can drive a coding agent. Such \emph{lightweight} models are attractive for privacy, cost and offline use, but summary-based compaction serves them poorly in two ways. First, a model-written summary is slow and fragile on local hardware. Second, and more fundamentally, a lighter model is less able to \emph{compensate} for a lossy summary: to notice what is missing, guess what an abbreviated line meant, and search its history persistently until it finds the original.

\paragraph{} We propose that compaction for lightweight models should \emph{guarantee transparency} in three senses:
\begin{enumerate}[label=(\roman*)]
  \item \textbf{What is kept in context is kept verbatim}: the user's and the agent's own words are retained absolutely, not as a reinterpretation.
  \item \textbf{What is removed is explicitly declared}, so the model knows what it no longer has.
  \item \textbf{All of what is removed is recoverable in one step}, and recovery protocols and conditions are parsimonious and well defined.
\end{enumerate}
\vb{} \citep{woolford2026transparent}, which we built as an extension for the open-source Pi coding agent \citep{zechner2026pi}, implements these properties (\cref{sec:method}). It removes tool outputs and reasoning, which make up most of the context and can be re-read or recovered, and keeps everything that was \emph{said} by both the user and agent. 
The guiding principle is to aid the model in avoiding confabulation and false certainty:
Summary and information compression invites misplaced certainty on partial information.
Conversely, declaring what information is removed tells the agent which of its beliefs rest on material it no longer has, keeping the original words leaves no paraphrase to misread, and one-step recovery lets the agent check instead of guessing.

In this study we compared \vb{} with Pi's default model-written summary (\pd) and with \vcc{} \citep{monotykamary2026pivcc}, a widely used deterministic extractor for Pi that is functionally closer to \vb{}, on two kinds of probe: commitments in real sessions, and facts planted in synthetic sessions and probed after repeated compaction. 
Transparency matters most for the kind of error that a lossy context invites. When information is missing and the agent knows it, the agent can look it up, check, or say that it does not know, and the user can step in: the failure is visible and cheap to repair. When the context instead holds something that only looks like the answer, such as a reworded rule, a stale value or a fragment of code, the agent acts on it with confidence, and the user has no reason to check. In a coding agent such confident errors turn directly into actions, such as a broken commitment or a wrong value written into a file. We therefore distinguish confident blunders separately from uncertain errors and reports of missing knowledge.
The main model is \qwen{} running locally. Two other lightweight models (\luna{} and \gemma) and a frontier model (\sol) test how the results depend on the model. Scoring was blind to the method, and a replication package is available publically.

\paragraph{Contributions:}
\begin{itemize}
  \item \textbf{A method:} \vb{}, a minimal, deterministic, verbatim compaction scheme with delegated recovery over the raw session history, released as a Pi extension (\cref{sec:method}).
  \item \textbf{Evidence on two probe sets} (\cref{fig:threeway}, \cref{sec:results}). On commitments, \vb{} led both alternatives on all three lightweight models, and \vcc{}'s failures were mostly confident. On planted facts it led \pd{} by 22--42 points.
  \item \textbf{A mechanism} (\cref{sec:mechanisms}). Both lossy methods \emph{select}. Extraction cuts the qualifying clauses that carry a rule; rewriting keeps rules but narrows or reframes them, and every such error was confident. \vb{} never lacked a commitment that another method kept.
  \item \textbf{Model dependence} (\cref{sec:models}). How well lossy compaction works depends on how hard the model searches. The same \vcc{} context gave 98\% recall on Qwen and 46\% on Luna. \vb{} held at 95--98\% on both, and its advantage on commitments was smaller on the frontier model.
\end{itemize}

\section{Background and Related Work}
\label{sec:related}
Keeping information outside a model's context and retrieving it only when it is needed is not a new
idea. Retrieval-augmented generation conditions a language model on passages fetched from an       
external store at inference time \citep{lewis2020rag}, and agent frameworks have since applied the 
same principle to the agent's own history and workspace.
When a conversation exceeds the context window, an agent can summarise it, drop parts of it, or move it to external storage that the model queries on demand. MemGPT treats the context window as a small working memory backed by larger storage, with the model paging information in and out \citep{packer2023memgpt}. Anthropic's guidance on context engineering describes compaction as summarising a conversation near the limit and restarting from the summary, and calls the clearing of old tool results ``one of the safest lightest touch forms of compaction'' \citep{anthropic2025context}. Longer windows are only a partial substitute, since models use information in the middle of a long context less reliably than information near its ends \citep{liu2024lost}.

The closest relative of our method is \emph{observation masking}. \citet{lindenbauer2025complexity} compared LLM summarisation with masking, which replaces older tool outputs with a placeholder, on SWE-bench Verified. With Qwen3-Coder 480B, masking solved about as many instances as summarisation (54.8\% against 53.8\%), and both roughly halved the cost per instance relative to an agent without context management (reductions of 52.7\% and 50.4\%). \vb{} extends masking in three ways: it masks the model's reasoning as well as tool outputs, it survives repeated compaction by carrying forward the prose of every earlier span, and it makes everything it masks recoverable. Where that work measured end-task success, we measure what the agent still knows and acts on after compaction.

Compaction itself has recently been evaluated more directly. Factory compared three production compressors using probes for recall, artifacts, continuation and decisions, scored by an LLM judge, including over repeated compression \citep{factory2025compression}. TRACE evaluates single compaction events through paired continuations from the same environment state \citep{min2026trace}. Closest to our commitment probes, \citet{chen2026governance} showed that compaction erases in-context governance constraints, and proposed \emph{constraint pinning}, which quarantines the constraints from lossy compaction. \vb{} pins every user message by construction, so it needs no step that decides which lines are rules. Even without compaction, prohibitions decay under context pressure while requirements persist \citep{gamage2026omission}.

\paragraph{}
Pi \citep{zechner2026pi} is an open-source, extensible coding agent. Its default compaction asks the model for a structured summary with sections for the goal, ``Constraints \& Preferences'', progress, ``Key Decisions'', next steps and critical context; later compactions update the previous summary. Extensions can replace compaction entirely.

 \vcc{} \citep{monotykamary2026pivcc}, inspired by the View-oriented Conversation Compiler \citep{zhang2026vcc}, builds its summary without model calls. It extracts the session goal, files and changes, commits, outstanding errors and user preferences into sections, keeps a brief transcript, and offers a search tool over the raw session for recovery. 
We chose \vcc{} as our main comparison because it is a widely used extension that closely resembles \vb{}: both replace Pi's compaction without any model calls, both produce checkpoints deterministically and both keep the full history recoverable through a search tool over the raw session. Their raw session recovery methods differ but have comparable costs in terms of time and output token usage. The key differences lie in what remains in the agent's context, and how the full history is structured for retrieval. \vb{} guarantees that text is either kept word for word or removed from the context entirely and replaced with a flag depending on type, whereas \vcc{} reduces the text algorithmically, extracting lines that match keywords and phrases into sections and truncating the rest to first lines, one-line turns and a capped transcript. Likewise, \vb{}'s retrieval is comparatively straightforward, giving the subagent basic tools to scan the raw session file, whereas \vcc{} utilizes more post-hoc mechanisms for structuring and parsing the history.

Two other popular extensions differ in ways that make \vcc{} the cleaner comparison. \textsf{pi-blackhole} \citep{k0valik2026blackhole} builds on \vcc{}'s compaction code but adds an observational memory written by background model calls, so it would not isolate the compaction step. \bil{} \citep{ranxianglei2026billion} belongs to a different category, self-compression, in which the model decides what to compress by calling a tool.

\section{Experimental Setup}
\label{sec:setup}
\subsection{\vb{}}
\label{sec:method}

\subsubsection{Checkpoint Retained in Context}

At each compaction, \vb{} replaces the older history with a checkpoint (\cref{lst:checkpoint}), and, as per Pi's standard behaviour, keeps the most recent part of the conversation unchanged after it (its \texttt{keepRecentTokens} setting). For the span being compacted, the checkpoint keeps every user message and all assistant prose word for word. It reduces each tool call to a one-line signature, consisting of the tool name and truncated arguments, so that the agent knows what it did, and it lists the files read and modified, each marked as not in context. Tool outputs and the model's reasoning are removed. Spans from earlier compactions are carried forward, so that after several compactions the checkpoint still holds everything that was said in the session.

The checkpoint also declares what it removed. Its header states what is kept and what is not, and that retained prose is ``a record of what was said, not verification of its claims''. A short re-orient block follows. It asks the agent to re-read the documents its instructions require, to check the state of the repository before relying on it, to re-read any file before editing it, and to treat anything it knows only from removed reasoning or tool output as unverified until checked.

\subsubsection{Recovery}

Pi's session file is append-only, so everything that compaction removed remains on disk. \vb{} exposes it through a single tool, \texttt{context\_lookup}. The agent asks a question in natural language, and a subagent searches the full session history, including earlier reasoning and complete tool outputs, in its own context. It returns only the relevant findings, and labels those that rest on the model's earlier reasoning. Three bounded fallback tools (list, grep and show entries) are available when the findings are incomplete, and consecutive lookups are capped. The delegation is deliberate: the main agent makes one call and receives an answer, instead of paging through search results itself. This reduces the risk of re-introducing excessive amounts of the removed history back into the compacted context.

\subsubsection{Budget}

As \vb{} avoids compressing text, it is not possible to maintain its guarantees through an indefinite number of compactions. The conversation section of the checkpoint is capped, by default at 25\% of the model's window. With realistic usage, this allows for approximately 10 rounds of compaction before capping begins to apply. Once this upper bound of verbatim compactions is reached, conservative truncation begins to be applied similarly to \vcc{}, with similar behaviour expected at maximal truncation ($>20$ rounds).
Testing behaviour after large numbers of compactions is beyond the scope of this study.

 \Cref{tab:methods} contrasts \vb{} with the two alternatives studied here.
\begin{table}[t]
\centering
\small
\begin{tabularx}{\linewidth}{>{\raggedright\arraybackslash}p{2.6cm}>{\raggedright\arraybackslash}X>{\raggedright\arraybackslash}X>{\raggedright\arraybackslash}X}
\toprule
 & \vb & \vcc & \pd \\
\midrule
Written by & code (deterministic) & code (deterministic) & the model \\
User messages & word for word & extracted: goal lines, pattern-matched preferences, one-line earlier turns, brief transcript & reworded into template sections \\
Assistant prose & word for word & one-line turns, brief transcript & reworded \\
Tool calls & one-line signatures, file lists & one-liners in the brief transcript, file and commit lists & summarised, file lists \\
Tool outputs, thinking & removed, declared & errors extracted, rest removed & summarised or dropped \\
Recovery & \texttt{context\_lookup} subagent (+ 3 fallback tools), recovery protocol appended in context & \texttt{vcc\_recall} search & none \\
Compaction time & 0.43\,s & 0.40\,s & 167.8\,s \\
\bottomrule
\end{tabularx}
\caption{\textbf{What each method keeps.} Compaction times are means per compaction with \qwen{} in the planted-fact study.}
\label{tab:methods}
\end{table}

\begin{listing}[t]
\begin{tcolorbox}[colback=panel,colframe=panel,boxrule=0pt,arc=2pt,left=6pt,right=6pt,top=4pt,bottom=4pt,fontupper=\footnotesize\ttfamily]
\#\# Verbatim compaction checkpoint (compaction 3 on this branch)\\
- In the compacted spans, user and assistant prose is kept verbatim [\dots]; tool calls are kept as one-line signatures (arguments truncated). Assistant thinking and tool outputs are removed. Retained prose is a record of what was said, not verification of its claims.\\
- The full transcript before this point is kept in the session; use `context\_lookup` to recover details (including thinking and tool outputs).\\[3pt]
**Re-orient before continuing.** [re-read required documents; check git status; re-read files before editing; treat pre-compaction knowledge from thinking or tool output as unverified]\\[3pt]
\#\#\# Compacted conversation (oldest first)\\
<span 1 \dots> [User]: \dots \quad [Assistant]: \dots \quad [tool] read(src/app.ts) \dots </span>\\
<span 2 \dots> \dots </span>\\[3pt]
<read-files note="read before compaction; contents NOT in context"> \dots </read-files>\\
Conversation from <time> onward continues verbatim below --- it was not compacted.
\end{tcolorbox}
\caption{\textbf{Structure of a \vb{} verbatim checkpoint} (abridged from the extension's source; bracketed text is summarised here). The original user and assistant text of every compacted span appears under ``Compacted conversation''.}
\label{lst:checkpoint}
\end{listing}


\subsection{Methods and Models}

We compare \vb{} (v0.8.0), \vcc{} (v0.8.12, see Appendix \ref{vcc-app} for discussion of different \vcc versions), and \pd (v1.1.0). In the \pd{} condition the summaries are written by the same model that answers the probes. All methods run inside Pi with the same model, settings, prompts and session material.

Four models were used. \qwen{} \citep{qwen38} (\texttt{unsloth/Qwen3.8-27B-GGUF}, 6-bit quantisation, xhigh effort) and \gemma{} \citep{gemmateam2026gemma4} (\texttt{unsloth/gemma-4-31B-it-GGUF}, 5-bit quantisation, reasoning enabled) run locally on llama.cpp. Qwen runs at its highest reasoning setting with temperature 0.7, top-$k$ 20 and top-$p$ 0.95; Gemma uses its recommended sampling. \luna{} and \sol{} run through the OpenAI Codex provider at medium reasoning, which exposes no temperature or seed. We class Qwen, Gemma and Luna as \emph{lightweight} models and Sol as a \emph{frontier} model. \vb's lookup subagent always uses the same model as the agent.

\subsection{Probe Sets}

\Cref{tab:probesets} summarises the two probe sets. \emph{Commitment} probes test whether the agent still honours what was settled in a real conversation. For each session, a fresh Claude Opus writer, briefed without any mention of compaction methods, identified commitments made early in the session: constraints (something the user required, forbade or waived) and decisions (a choice or convention that was agreed). For each commitment it wrote a natural next user message that would quietly break the commitment if the agent had forgotten it, without mentioning it. A separate Opus validator checked each probe against eight criteria. The commitment's key strings had to appear exactly once in the session, so that nothing later restates it, and each probe has a control message that states the commitment; probes whose control failed were excluded, leaving 27 probes in 18 sessions.

\emph{Planted-fact} probes test recall after repeated compaction. Eighteen synthetic debugging sessions, six stories with distinct plots each instantiated with three sets of values, are built to resemble real sessions, with about 85\% of their tokens in tool output. Seventeen details are planted early in each session, among them codes seen once in a log, values that change twice, details that appear only in tool output or only in the model's reasoning, and asides about hosts and people. Each session is compacted three times, each time at about 95k tokens, as for a 128k window, and after each compaction the agent answers questions on all planted details and on three lures. Every round opens with an episode that creates look-alikes of the planted values, so that a vague memory does not suffice.

The real sessions come from a public dataset of Pi sessions \citep{adeola2026traces}, restricted to MIT or CC-BY-4.0 licences. Each session is replayed through each method's actual compaction at a 48k-token window, so long sessions are compacted many times before the probe. No session text, probe text or model reply is published.

\begin{table}[t]
\centering
\small
\begin{tabularx}{\linewidth}{>{\raggedright\arraybackslash}p{2.6cm}>{\raggedright\arraybackslash}X>{\raggedright\arraybackslash}X}
\toprule
 & Commitments & Planted facts \\
\midrule
Material & 18 real sessions & 18 synthetic sessions \\
Compaction & replay at 48k, many compactions & 3 compactions at $\sim$95k tokens \\
Probes & 27 (14 constraints, 13 decisions) & 17 facts + 3 lures after each compaction \\
Scoring & blind LLM judge; failures labelled confident or not & by rule \\
Models (reps) & Qwen, Luna, Gemma, Sol (3; 5 in the first Qwen run) & Qwen, Luna, Gemma, Sol (1) \\
\bottomrule
\end{tabularx}
\caption{\textbf{Probe sets.} ``Reps'': independent replies per probe and method.}
\label{tab:probesets}
\end{table}

\subsection{Isolation and Scoring}

The agent must not reach what compaction removed except through its method's own recovery tool. Probe turns therefore run on a copy of the compacted session with no file, shell or edit tools, and the only tool available is the method's recovery tool (\texttt{context\_lookup} or \texttt{vcc\_recall}; \pd{} has none). In the commitment study, an arm-neutral sentence tells the agent that it has no shell or file access in this turn. Probe turns declare a 128k window, so that they cannot trigger a further compaction.

Planted-fact answers are scored by rule, and a \emph{confident wrong} answer is a wrong or superseded value that is not marked as unverified or unknown. Commitment replies are scored by blind Claude Sonnet judges, who see the commitment, the instruction, the pass criterion and the reply, but not the method, the session or the repetition. On commitment probes the judge also labels each failure as \emph{confident}, meaning that the reply went ahead without any doubt, question or flag, or not. A second blind judge re-scored a seeded 25\% of the replies in each run, with agreement between $\kappa = 0.67$ and $\kappa = 1.0$.

\subsection{Analysis}

The planted-fact and commitment studies are analysed with two-way repeated-measures ANOVAs with the factors model and method. For planted facts the subjects are the 18 sessions, which are identical on every model, and the outcome is each session's recall over the three compactions, on Qwen, Luna and Gemma. For commitments the subjects are the 27 probes and the outcome is each probe's pass rate over three repetitions, on all four models; failed turns, and the one probe whose session \pd{} could not replay on Qwen, count as failures. The first commitment run, on Qwen, used five repetitions per probe. We found three repetitions enough to reduce the noise from the model's sampling, and five excessive; because the runs were costly, all later runs used three, and the ANOVA uses repetitions one to three of the Qwen run. We apply Greenhouse--Geisser corrections and report partial $\eta^2$. To locate a method effect, we also fit the same ANOVA to each pair of methods, whose interaction term then tests whether that pair's gap differs between models. Because the outcomes are bounded rates, the $F$ tests are approximate. The planted-fact study on Sol was added later with two methods only; it is reported as an exploratory update, analysed with a paired sign-flip permutation test over sessions, and is not part of the ANOVA. Non-significant differences are reported with their estimate and $p$-value, as weak evidence in their direction rather than as evidence of no difference.

A replication package that re-runs the planted-fact and commitment studies with one's own model, Pi installation and judge is available at \url{https://github.com/felixwoolford/compaction-study-replication}.

\section{Results}
\label{sec:results}
\begin{figure}[t]
	\centering
	\includegraphics[width=\linewidth]{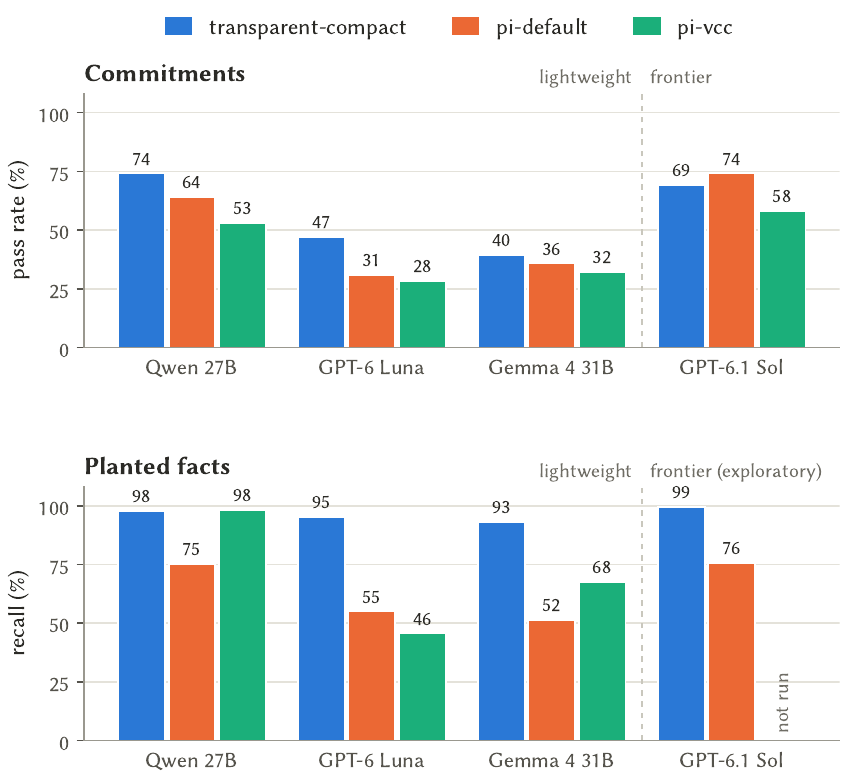}
	\caption{\textbf{The three methods on both probe sets.} \emph{Commitments}: 27 real-session probes $\times$ 3 repetitions on each model; Sol is a frontier model, the others are lightweight. \emph{Planted facts}: 18 synthetic sessions, each probed after three compactions; the Sol group is an exploratory update with two methods and is not part of the ANOVA. \vcc{} was not run on Sol because of cost-- extrapolating from earlier data we expect that it would approximately match \vb{}'s near perfect performance with that model. Statistics are in \cref{tab:anova}.}
	\label{fig:threeway}
\end{figure}

\Cref{fig:threeway} shows the three methods on both probe sets, and \cref{tab:anova} gives the ANOVAs.

\begin{table}[t]
\centering
\small
\begin{tabular}{@{}llrrr@{}}
\toprule
 & Effect & $F$ (df) & $p$ & $\eta_p^2$ \\
\midrule
\multicolumn{5}{@{}l}{\emph{Commitments} (27 probes; Qwen, Luna, Gemma, Sol)}\\
\quad Omnibus & model & 15.12 (3, 78) & $<$0.001 & 0.37 \\
 & method & 5.37 (2, 52) & 0.008 & 0.17 \\
 & model $\times$ method & 1.43 (6, 156) & 0.22 & 0.05 \\
\quad \vb{} vs \vcc & method & 9.98 (1, 26) & 0.004 & 0.28 \\
 & model $\times$ method & 1.35 (3, 78) & 0.27 & 0.05 \\
\quad \vb{} vs \pd & method & 2.25 (1, 26) & 0.15 & 0.08 \\
 & model $\times$ method & 1.82 (3, 78) & 0.16 & 0.07 \\
\quad \pd{} vs \vcc & method & 3.28 (1, 26) & 0.08 & 0.11 \\
 & model $\times$ method & 1.05 (3, 78) & 0.37 & 0.04 \\
\addlinespace
\multicolumn{5}{@{}l}{\emph{Planted facts} (18 sessions; Qwen, Luna, Gemma)}\\
\quad Omnibus & model & 164.6 (2, 34) & $<$0.001 & 0.91 \\
 & method & 317.0 (2, 34) & $<$0.001 & 0.95 \\
 & model $\times$ method & 62.0 (4, 68) & $<$0.001 & 0.79 \\
\quad \vb{} vs \pd & method & 860.2 (1, 17) & $<$0.001 & 0.98 \\
 & model $\times$ method & 42.1 (2, 34) & $<$0.001 & 0.71 \\
\quad \vb{} vs \vcc & method & 324.6 (1, 17) & $<$0.001 & 0.95 \\
 & model $\times$ method & 84.5 (2, 34) & $<$0.001 & 0.83 \\
\quad \pd{} vs \vcc & method & 35.0 (1, 17) & $<$0.001 & 0.67 \\
 & model $\times$ method & 44.7 (2, 34) & $<$0.001 & 0.72 \\
\bottomrule
\end{tabular}
\caption{\textbf{Model $\times$ method repeated-measures ANOVAs.} The omnibus test uses all three methods; each pairwise ANOVA uses one pair of methods, and its interaction tests whether that pair's gap differs between models. $p$-values are Greenhouse--Geisser corrected.}
\label{tab:anova}
\end{table}

\subsection{Commitments}
\label{sec:commitments}

Across the four models, the compaction method had a reliable effect on how often the agent honoured a commitment ($F(2,52) = 5.37$, $p = 0.008$, $\eta_p^2 = 0.17$). The models also differed strongly in overall level ($F(3,78) = 15.1$, $p < 0.001$): Qwen and Sol passed most probes, Luna and Gemma far fewer (\cref{tab:s4}). The model $\times$ method interaction was not significant ($F(6,156) = 1.43$, $p = 0.22$), so the ordering of the methods did not differ reliably between models.

The pairwise ANOVAs locate the method effect. \vb{} outperformed \vcc{} ($F(1,26) = 9.98$, $p = 0.004$, $\eta_p^2 = 0.28$), and did so on every model: by 21 points on Qwen, 19 on Luna, 7 on Gemma and 11 on Sol, with no reliable difference in the size of the gap between models ($F(3,78) = 1.35$, $p = 0.27$). On Qwen, where five repetitions were run, comparison with \vcc{} gave 75.6\% against 54.8\% ($p = 0.001$), and it held in every sensitivity analysis.

Against \pd{} the difference was smaller and not significant overall ($F(1,26) = 2.25$, $p = 0.15$, $\eta_p^2 = 0.08$). Its pattern across models is nonetheless the one our argument predicts. \vb{} was ahead on the three lightweight models, by 10 points on Qwen, 16 on Luna and 4 on Gemma, and behind by 5 on the frontier model, where Pi's own summary had the highest pass rate of any condition (74.1\% against 69.1\%). The corresponding interaction did not reach significance ($F(3,78) = 1.82$, $p = 0.16$), so this is a pattern rather than a demonstrated interaction. \pd{} in turn tended to outperform \vcc{} ($F(1,26) = 3.28$, $p = 0.08$).

\begin{table}[t]
\centering
\small
\rowcolors{3}{rowshade}{}
\begin{tabular}{@{}lrrrcrrr@{}}
\toprule
 & \multicolumn{3}{c}{Pass rate} && \multicolumn{3}{c}{Failed confidently} \\
\cmidrule(lr){2-4}\cmidrule(l){6-8}
Model & \vbh & \pd & \vcc && \vbh & \pd & \vcc \\
\midrule
\qwen{} & \textbf{74.1} & 64.2 & 53.1 && \textbf{14.8} & 22.2 & 40.7 \\
\luna{} & \textbf{46.9} & 30.9 & 28.4 && \textbf{34.6} & 46.9 & 51.9 \\
Gemma 4 31B & \textbf{39.5} & 35.8 & 32.1 && \textbf{37.0} & 48.1 & 55.6 \\
\sol{} (frontier) & 69.1 & \textbf{74.1} & 58.0 && \textbf{23.5} & \textbf{23.5} & 33.3 \\
\bottomrule
\end{tabular}
\caption{\textbf{Commitments: pass rate and confident failures} (\%, 27 probes $\times$ 3 repetitions per model; failed turns and missing cells count as failures). Pass rates are the ANOVA's cell means.}
\label{tab:s4}
\end{table}

\begin{figure}[t]
  \centering
  \includegraphics[width=\linewidth]{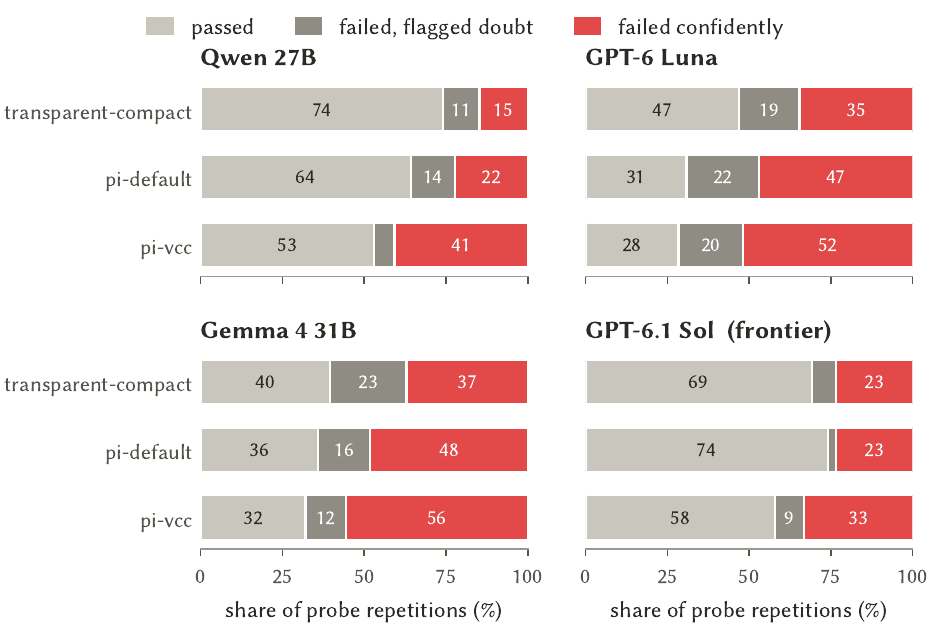}
  \caption{\textbf{How commitment probes failed.} Each bar divides all probe repetitions into passes, failures that flagged doubt (a question, warning or hedge) and confident failures; 27 probes $\times$ 3 repetitions per model.}
  \label{fig:outcomes}
\end{figure}

How the agent failed matters as much as how often. A commitment broken without any sign of doubt gives the user no reason to check, whereas a reply that raises a question or flags a conflict does. \Cref{fig:outcomes} divides every probe repetition into passes, failures that flagged doubt and confident failures. With \vcc{}, between a third and over half of all replies broke the commitment confidently (41\% on Qwen, 52\% on Luna, 56\% on Gemma and 33\% on Sol), against 15--37\% with \vb{}. Pi's summary fell between the two on the lightweight models and matched \vb{} on Sol. In the per-model tests, \vb{} produced fewer confident failures than \vcc{} on Qwen ($p < 0.001$) and Luna ($p = 0.008$), with the same direction on Gemma ($p = 0.052$) and Sol ($p = 0.12$).

Gemma passed the fewest probes with every method. It rarely used its recovery tools: with \vb{} it made a lookup in 12 of 79 turns, and on one probe it repeated identical searches until the turn timed out.

\subsection{Planted Facts across Repeated Compaction}
\label{sec:facts}

\vb{} recalled more planted facts than Pi's default summary on every model (\cref{tab:s1}): 97.8\% against 75.3\% on Qwen, 95.1\% against 55.0\% on Luna and 93.3\% against 51.5\% on Gemma. This is the largest and most consistent effect in the study ($F(1,17) = 860.2$, $p < 0.001$, $\eta_p^2 = 0.98$ in the pairwise ANOVA). \vb{} also gave fewer confident wrong answers than \pd{} on all three models, by 3.5 points on Qwen, 5.4 on Luna and 4.4 on Gemma (Holm-corrected $p \le 0.006$ in each case). The gap in recall was 22.5 points on Qwen and nearly twice that on Luna (40.1) and Gemma (41.7), so the pair's interaction with the model was significant as well ($F(2,34) = 42.1$, $p < 0.001$).

\vcc{} behaved differently. It matched \vb{} on Qwen (98.2\% against 97.8\%) but fell to 46\% on Luna and 68\% on Gemma, where its recall declined with each compaction (73.5\%, 69.0\% and 60.5\%). \vb{}'s recall, by contrast, held between 93\% and 98\% on all three models. The strong model $\times$ method interaction of the omnibus ANOVA (\cref{tab:anova}: $F(4,68) = 62.0$, $p < 0.001$) therefore reflects how the two lossy methods fared on each model rather than any change in \vb{}; \cref{sec:models} traces \vcc{}'s part of it to how persistently each model searched.

As an exploratory update, the planted-fact study was later also run on the frontier model, Sol, with \vb{} and \pd{} only (\cref{tab:s1}). \vb{} recalled 99.3\% against 75.6\% for Pi's summary, more in all 18 sessions ($+23.7$ points [20.6, 26.9], $p < 0.001$), and gave no confident wrong answers, against 2.1\% for \pd{} ($p = 0.002$). Pi's summary thus lost about as much on the frontier model as on Qwen (75.6\% against 75.3\%), although on Sol most of its losses were abstentions: of its 918 answers, 190 were ``unknown'' and 34 were wrong.

\begin{table}[t]
\centering
\small
\begin{tabular}{@{}lrrr@{}}
\toprule
Model & \vb & \pd & \vcc \\
\midrule
\qwen{} & 97.8 & 75.3 & \textbf{98.2} \\
\luna{} & \textbf{95.1} & 55.0 & 45.6 \\
Gemma 4 31B & \textbf{93.3} & 51.5 & 67.5 \\
\sol{} (frontier)$^\dagger$ & \textbf{99.3} & 75.6 & --- \\
\bottomrule
\end{tabular}
\caption{\textbf{Planted facts: recall} (\%, 18 sessions, three compactions). Qwen, Luna and Gemma are the ANOVA's cell means; Sol$^\dagger$ is an exploratory update with two methods and is not part of the ANOVA.}
\label{tab:s1}
\end{table}

Twelve of the Qwen sessions were continued to eight compactions. \vb{}'s recall stayed between 96.6\% and 100\% in every round, whereas Pi's summary failed in most sessions at the fifth or sixth compaction, when the growing summary hit its generation cap. Over rounds four to eight, \vb{} gave 6.5 points fewer confident wrong answers ($p = 0.006$).

\subsection{Speed, Token Cost and Robustness}
\label{sec:robustness}

With Qwen, \vb{} took 0.43\,s per compaction and \vcc{} 0.40\,s, whereas \pd{} took 167.8\,s because the local model writes the summary with full reasoning. Sol's summaries took about 61\,s each for \pd{}.

\vb{}'s context is larger, and its lookups add tokens (\cref{tab:cost}). On Qwen's commitment probes, its probe turns processed on average 57{,}300 prompt tokens that were not served from the cache, against 21{,}389 for \vcc{} (2.7 times as many); on Gemma the ratio was 1.7. Output tokens were similar. In the planted-fact sessions, where every probe turn required a lookup, \vb{}'s sessions took about three minutes longer in total than those of \pd{} despite its near-instant compaction (14.1 against 11.2 minutes, $p = 0.077$), and had no significant difference in duration compared to \vcc{} ($p = 0.78$). It should be noted that as these sessions were entirely based on recalling information from earliest in the interaction, agents performed significantly more context retrievals than usual, and thus retrieval time dominated total session time in a way that would not be expected in practical use. 

A model-written summary can also fail outright. At Pi's stock settings for a 48k window, the summary may use 0.8 times the 8{,}192-token reserve, or 6{,}553 tokens; Qwen at its highest reasoning setting spent that budget thinking, and compaction failed in 15 of 16 real sessions. To allow for comparison with the other methods, we allowed a larger summary budget of the planted-fact study (26{,}214 tokens). Subsequently, 202 of 205 compactions succeeded, but 3 of 41 long sessions still failed after 7--9 compactions, when the repeatedly updated summary outgrew the budget. All \pd{} results on Qwen in this paper use the larger budget, which favours \pd{} relative to its stock behaviour. The summaries written by Luna, Gemma and Sol did not fail. Deterministic compaction cannot fail in this way.

\begin{table}[t]
\centering
\small
\rowcolors{3}{rowshade}{}
\begin{tabular}{@{}lrrrr@{}}
\toprule
 & \multicolumn{2}{c}{Qwen} & \multicolumn{2}{c}{Gemma} \\
\cmidrule(lr){2-3}\cmidrule(l){4-5}
 & \vbh & \vcc & \vbh & \vcc \\
\midrule
Prompt tokens & 57{,}300 & 21{,}389 & 34{,}576 & 20{,}822 \\
Output tokens & 5{,}975 & 5{,}656 & 1{,}419 & 1{,}440 \\
Recovery calls (total) & 186 & 1{,}238 & 27 & 65 \\
Turns & 139 & 139 & 79 & 80 \\
\bottomrule
\end{tabular}
\caption{\textbf{Token cost per commitment probe turn} (\vb{} against \vcc; local models, measured at the server). ``Prompt tokens'': mean prompt tokens processed per turn, excluding cache hits and including \vb{}'s lookup subagent.}
\label{tab:cost}
\end{table}

\section{Discussion}
\subsection{Why the Alternatives Underperform}
\label{sec:mechanisms}

The two clearest method effects run in opposite directions for the two alternatives. Pi's summary fell far behind \vb{} on planted facts but not significantly on commitments, whereas \vcc{} fell behind it on commitments but matched it on planted facts when the model searched persistently. Failure analyses of both studies, summarised in this section, trace each pattern to what the method chooses to keep. Both analyses are exploratory. The planted-fact analysis uses all 18 Qwen sessions for recall by source and the 12 sessions continued past three compactions for the review of confusions; the commitment analysis covers every Qwen probe under all three methods (26 probes $\times$ 3 repetitions $\times$ 3 methods).

\subsubsection{Pi's Summary Loses Details and the Context That Separates Them}

A model-written summary keeps what the user said and compresses the rest. On Qwen's planted facts, \pd{} recalled 97.9\% of the details the user had stated, 80.2\% of those in assistant prose, 62.7\% of those seen only in tool output and 15.8\% of those that existed only in the model's reasoning. Because \pd{} has no recovery tool, whatever the summary omits is lost for good. \vb{} removes tool outputs and reasoning from the context as well, but its agent recovered them through \texttt{context\_lookup} and reached 100\%, 100\%, 97.6\% and 87.4\% for the same four sources. The lookup was used in every probe turn, because the questions were built to require recovery, so this rate does not reflect ordinary use.

Omission is not the only failure. In each of the 7 cases in which a detail was first answered correctly and later answered confidently wrong, the correct value was still in the summary. In 5 of them a competing value from a later look-alike episode stood beside it, and the context that had separated the two was gone. In one session, for example, the summary recorded that a backfill job's batch size had been changed to 7{,}500 after benchmarks; after a later episode about a different job it also recorded ``Chose 2750 because it runs clean'', no longer saying which job that referred to, and the agent then gave 2{,}750 as the agreed batch size. One summary also silently rewrote the day of a release freeze, an error the probes did not score, so they understate summary errors. The synthetic sessions were built to contain such look-alikes, so the rate of confusion reflects the material, but the mechanism does not: a summary can keep a value while losing the context that binds it to its episode, whereas \vb{} keeps the original messages in which that context lives.

\subsubsection{\vcc{} Has No Place for Rules}

\Cref{tab:presence} shows where each commitment stood in each method's context. \vcc{} has no reliable place for rules. Its only rule-like section, ``User Preferences'', is filled by English-only patterns such as ``prefer\,\dots'' and ``never use\,\dots'', and it appeared in only 1 of 15 summaries, with an irrelevant line. Its other sections truncate: the goal section keeps the first line of recent user messages, earlier turns are cut to one short line each, and the transcript excerpt keeps only its last 120 lines. These cuts often fell just before the qualifying clause that carried the constraint, and procedural rules survived only as traces of what had been done (a wait call, a worktree removal) rather than as the rule behind them. Language was not the main cause: 11 of the 15 sessions were mainly in English, and \vcc{}'s pass rate was about the same in those sessions (53\%) as in the others (50\%).

\vcc{} did search its history, calling its recall tool in 64 of 78 runs, and 30 of its 37 failures came after a search. The searches, however, looked for code to act on rather than for what had been agreed, and the retrieved code made the replies concrete and therefore confident: 19 of its 33 confident failures opened by saying that the exact code had been found in the history. Recovery by search works when the model knows what to look for, as with a planted fact named in the question; it does not help when the model does not know that a rule exists.

\begin{table}[t]
\centering
\small
\begin{tabularx}{\linewidth}{@{}>{\raggedright\arraybackslash}p{3.6cm}>{\raggedright\arraybackslash}X>{\raggedright\arraybackslash}X>{\raggedright\arraybackslash}X@{}}
\toprule
 & \vb & \pd & \vcc \\
\midrule
Commitment present & 22 of 26, word for word & 13 of 26 in ``Constraints \& Preferences'' or ``Key Decisions'', plus 3 in the goal or kept tail & 9 of 26 intact; 7 truncated or only a trace of an action; 10 absent \\
Pass rate when present & 54/66 & 31/39 (in its constraint sections) & 19/27 (when intact) \\
Median summary size & 31k chars & 34k chars & 9.5k chars \\
\midrule
\multicolumn{4}{@{}l}{\emph{On the 21 probes left after removing 4 that no method kept and 1 flawed probe}}\\
Pass rate & 83\% & 78\% & 62\% \\
Confident failures & 6 & 10 & 22 \\
\bottomrule
\end{tabularx}
\caption{\textbf{Where the commitment stood in each method's context} (Qwen, 26 commitment probes; case review).}
\label{tab:presence}
\end{table}

\subsubsection{Pi's Summary Keeps Rules but Rewords Them}

Pi's summary template, by contrast, requires ``Constraints \& Preferences'' and ``Key Decisions'' sections, and Qwen filled them with the user's rules, often quoting them. Where a rule was present, \pd{} passed 31 of 39 times, with one confident failure. This is why it came close to \vb{} on commitments while falling far behind it on planted facts: the template keeps rules and decisions and compresses details such as values and outputs. Its commitment failures came from rewording rather than omission, since no \pd{} failure lost a commitment that another method's context kept. Instead the summary narrowed a rule's scope, recorded a standing rule as a fact about one past task, kept a decision from before the user reversed it, or recorded a rule only as a past change, and all nine such failures were confident. How well a summary keeps rules depends on the model that writes it, which is consistent with Sol's summaries giving the best commitment result of any condition (\cref{sec:commitments}).

\subsubsection{\vb{}'s Failures and advantages}
\label{sec:account}

\vb{} never lacked a commitment that another method had. Its 11 commitment failures with the rule present were misapplications, in which the agent saw or even quoted the rule and then went along with the request. \pd{} outperformed it on five probes; four of these differences amounted to a single repetition, and the fifth was a probe whose rule no method had, where \pd{}'s sensible default happened to pass. On four probes the commitment was in no method's context, because it lived only in a document read through a tool or was never stated as a rule, and all three methods failed most of these. Its remaining errors therefore lie in how the agent uses its context, not in what the checkpoint kept.

\paragraph{}

Both lossy methods \emph{select}, and each selection rule has a blind spot (\cref{tab:account}). Rewriting selects by the summariser's judgement. Pi's template asks for rules and decisions, so it keeps them, but it compresses the values, outputs and reasoning that planted-fact questions ask for, and loses the context that tells look-alike values apart; without a recovery tool, nothing it drops can be restored. Extraction selects by pattern and position. \vcc{} keeps a searchable record of everything, so facts can be recovered by a model that searches persistently, but it has no reliable place for a rule stated in ordinary language, and a model that does not know a rule exists does not search for it. In both cases the errors tend to be confident, because the agent's context holds something that looks like the answer: a stale value beside the right one, a reworded rule, or code retrieved from the history.

Each alternative is therefore competitive where its selection rule anticipates what is needed and falls behind where it does not, and the ANOVAs show exactly this. \vb{}'s significant advantages are over Pi's summary on planted facts ($F(1,17) = 860.2$, $p < 0.001$) and over \vcc{} on commitments ($F(1,26) = 9.98$, $p = 0.004$). Conversely, Pi's summary came within a non-significant margin of \vb{} on commitments, and \vcc{} matched it on planted facts with Qwen. \vb{} does not select what was said, and makes everything else recoverable in one delegated step, so it has no comparable blind spot: it is the only one of the three methods that did not fall significantly behind another method on any probe set. Since an agent cannot know in advance which kind of information a later turn will depend on, this robustness across kinds of information, rather than a win on any single probe set, is the main advantage of the method. It matters most for lightweight models, which are least able to reason or search their way around a blind spot (\cref{sec:models}).

\begin{table}[t]
\centering
\small
\begin{tabularx}{\linewidth}{@{}>{\raggedright\arraybackslash}p{1.7cm}>{\raggedright\arraybackslash}X>{\raggedright\arraybackslash}X>{\raggedright\arraybackslash}X>{\raggedright\arraybackslash}p{2.2cm}@{}}
\toprule
 & Values from tool output and reasoning & User rules and decisions & Recovery of what was dropped & Significantly behind \vb{} on \\
\midrule
\vb & removed but declared; recovered by lookup & kept word for word & one delegated lookup & no probe set \\
\pd & compressed or dropped; look-alikes confused & kept in template sections, sometimes reworded & none & planted facts \\
\vcc & removed; recoverable by search & truncated, a trace, or absent & search; depends on the model's persistence & commitments; planted facts on Luna and Gemma \\
\bottomrule
\end{tabularx}
\caption{\textbf{What each method keeps of the information the probes need}, summarising \cref{sec:mechanisms}.}
\label{tab:account}
\end{table}

\subsection{Model Dependence: Lightweight versus Frontier}
\label{sec:models}

\begin{figure}[t]
  \centering
  \includegraphics[width=\linewidth]{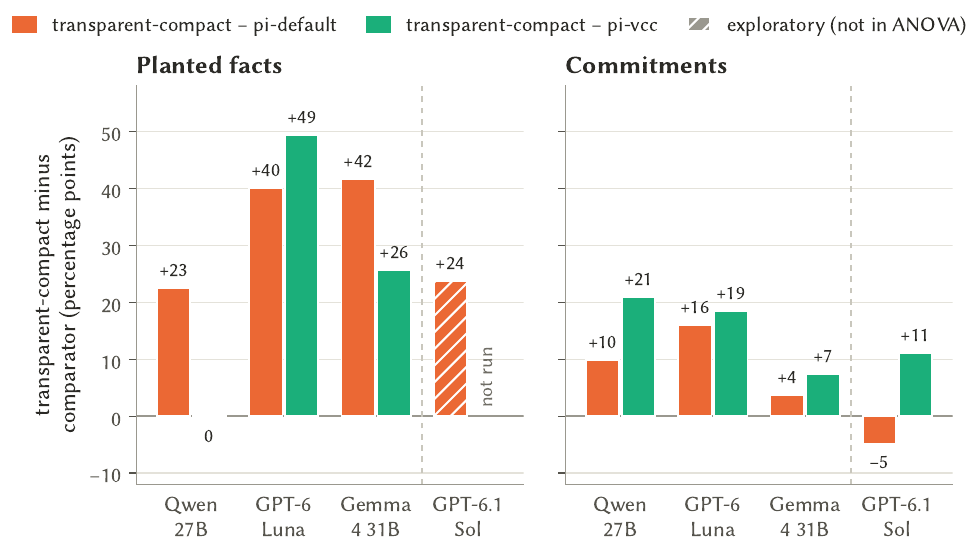}
  \caption{\textbf{\vb{}'s lead over each alternative, by model} (percentage points; differences between the ANOVA's cell means). Planted facts: 18 sessions; commitments: 27 probes $\times$ 3 repetitions. The dashed line separates the frontier model. The hatched Sol bar for planted facts is the exploratory update and is not part of the ANOVA. }
  \label{fig:gaps}
\end{figure}

\Cref{fig:gaps} sets \vb{}'s lead over each alternative side by side for every model, which is what the model $\times$ method interactions of \cref{tab:anova} test. Three observations follow.

First, how well lossy compaction works depends on how hard the model searches. \vcc{} gives an identical compacted context to every model on the planted-fact material, because its compaction is deterministic and the material is the same, yet its recall was 98\% on Qwen, 46\% on Luna and 68\% on Gemma. Qwen kept paging through \vcc{}'s search results until it found the values, with 30 or more recall calls in most probe batches. Luna made 5--14 calls per batch and then answered ``unknown'': 401 of its 918 planted-fact answers with \vcc{} were abstentions, against 4 for Qwen. The tool worked in both cases; what differed was how persistently the model used it. \vb{}'s lookup is a single delegated call, and its recall held between 93\% and 98\% on all three models.

Second, on commitments \vb{} barely needs to search. A commitment is something that the user or the agent said, so with \vb{} it is usually already in context. On the commitment probes it made 9 recovery calls in 81 turns on Luna (\vcc: 110), 15 on Sol (\vcc: 57) and 27 on Gemma (\vcc: 65). Its advantage on commitments therefore does not rest on the model's skill at searching. \vcc{}, by contrast, requires the model to realise that a rule exists, to search for it and to recognise it in the results, which on these probes rarely happened.

Third, the advantage depends on the model's strength. On the lightweight models \vb{} led both alternatives on commitments, although its lead over \pd{} was not significant overall. On the frontier model it still led \vcc{}, by 11 points, but Pi's own summary did slightly better than either (74.1\% against 69.1\%), consistent with a stronger model writing a summary that keeps the rules and making good use of it. On planted facts, likewise, both lossy methods fared best on Qwen. As a descriptive comparison only, Qwen with \vb{} passed as many commitment probes (74.1\%) as Sol did in its best condition.

We therefore claim that a lightweight model cannot be relied on to make up for lossy compaction, whether by reasoning or by searching, and that \vb{} removes that dependence. We do not claim that lightweight models fail with \vcc{} in general: on Qwen it matched \vb{} on planted facts.

\subsection{Practical and Theoretical Consequences}
\label{sec:discussion}

The account in \cref{sec:account} points to a design principle rather than a tuning problem. Any rule for choosing what to keep encodes a guess about what later turns will need, and we expect a better template or a better extractor to move the blind spot rather than remove it (See Appendix \ref{vcc-app} for an illustration of this in practice.). Commitments are a telling case: they are short, easy to miss, expressed in ordinary language rather than in any fixed form, and invisible in the environment. \citet{chen2026governance} found the same with compaction and safety policies: violations were 0\% when the constraint survived the summary and 38\% when it was dropped. Their remedy, pinning the constraints, requires knowing which lines are constraints. \vb{} keeps everything that was said, so nothing depends on recognising a rule, in any language. The cost of not selecting is a larger checkpoint, but because tool output dominates real sessions that cost is moderate: in our real sessions the median checkpoint (31k characters) was smaller than Pi's summary (34k).

The pattern has a long history in the psychology of memory. People quickly lose the exact wording of what they hear but keep its meaning \citep{sachs1967recognition}. Fuzzy-trace theory holds that a \emph{verbatim} trace and a \emph{gist} trace are stored in parallel, that the verbatim trace fades faster, and that gist-based memories support confident false recognition \citep{reyna1995fuzzy}. Bartlett showed that remembering is reconstructive, bending material toward the rememberer's schema \citep{bartlett1932remembering}, and people often misattribute the source of a memory \citep{johnson1993source}. A model-written summary is a gist trace produced on purpose. The failures of \pd{} resemble schema-driven reconstruction: rules were narrowed, or turned into facts about one past event, to fit the categories of a template, and every such failure was confident. \vb{} keeps the verbatim trace together with its source, recording who said what, and labels what came from the agent's own earlier reasoning as unverified. We offer this as an analogy that may guide design, not as evidence about how models work internally.

For practitioners, \vb{} particularly suits local and lightweight models, and sessions in which users state rules and preferences. It is also at least as good with the tested frontier model as the other methods, and avoids slow model-written compaction, but further testing would be required to endorse adoption exclusively for use with frontier models.  \vcc{} uses fewer prompt tokens per turn and recalls facts as well as \vb{} when the model searches persistently. Pi's default summary kept commitments best with the frontier model and reasonably well with Qwen, but it loses detail and, with weaker models, distorts rules confidently. Although the benchmarks for commitments use real session data, the probes do not necessarily reflect real world usage of agents, and \vcc{}'s maturity may make it more suitable in an applied setting. Nevertheless lthough the benchmark performance and simplicity of \vb{} invite further development of this method as an applied tool, and adoption of its principles.

\subsection{Limitations}                
\label{sec:limits}

The author built \vb{}. To limit the resulting bias, third-party methods were run as shipped with their defaults, probes were written by models briefed without any mention of the methods, scoring was blind, and a replication package is provided. 

The probes test what the agent knows and does in the next turn, with file and shell tools removed in order to isolate its memory of the conversation; real agents can also re-read files, which helps every method with facts but not with commitments.
 
Probes were written and validated by Claude Opus and scored by Claude Sonnet, with judge agreement between $\kappa = 0.67$ and 1.0. The models differ in provider and sampling as well as in capability: the API models allow no temperature or seed control, and control turns for the commitment probes were run on Qwen only (and on Sol for \vcc{} and \vb{}), so the model effect in the ANOVAs mixes these factors. The ANOVAs treat bounded rates as continuous, so their $F$ tests are approximate, and on Gemma one probe's timeouts count as failures.   
\section{Conclusion}
\label{sec:conclusion}

Compaction decides what a coding agent still knows. We compared three ways of compacting with four models, three of them lightweight. Keeping everything that was said word for word, dropping the bulky tool output and making the rest recoverable in one step was the only approach without a blind spot. It recalled far more planted facts than Pi's model-written summary on every model, kept the user's commitments better than \vcc{}'s extraction on every model, and did not fall significantly behind either alternative on any probe set. Each alternative fell behind where its rule for choosing what to keep failed to anticipate what was needed. Its lead over Pi's summary on commitments was confined to the lightweight models, and on the frontier model Pi's own summary did slightly better, though not significantly. This is consistent with our argument: simple, transparent compaction removes the need for the model to reason or search its way around what compaction lost, and lightweight models are the ones that most need that help.

\section*{Contributions and use of AI}
FW designed the studies, made all design and reporting decisions, and takes full responsibility for the content. FW drafted the manuscript and analysed experiments with assistance from Claude (Anthropic). To avoid author bias, Claude wrote and validated the probes and performed blind scoring (as described in \cref{sec:setup}) without FW's input. 

\section*{Code and data availability}
The \vb{} extension is available at \url{https://github.com/felixwoolford/transparent-compact-pi}. A replication package for the planted-fact and commitment studies, is available at \url{https://github.com/felixwoolford/compaction-study-replication}.

\bibliographystyle{plainnat}
\bibliography{refs}

\appendix
\section{\vcc{} Versions}
\label{vcc-app}

There are two popular and actively maintained versions of \vcc{} in the Pi package catalogue. @sting8k/pi-vcc is the original, and @monotykamary/pi-vcc forked from it in May 2026.
The two versions still maintain the same compaction core, and primarily differ in their retrieval processes. Neither is simply a superset of the other.

\begin{figure}
	\centering
	\includegraphics[width=\linewidth]{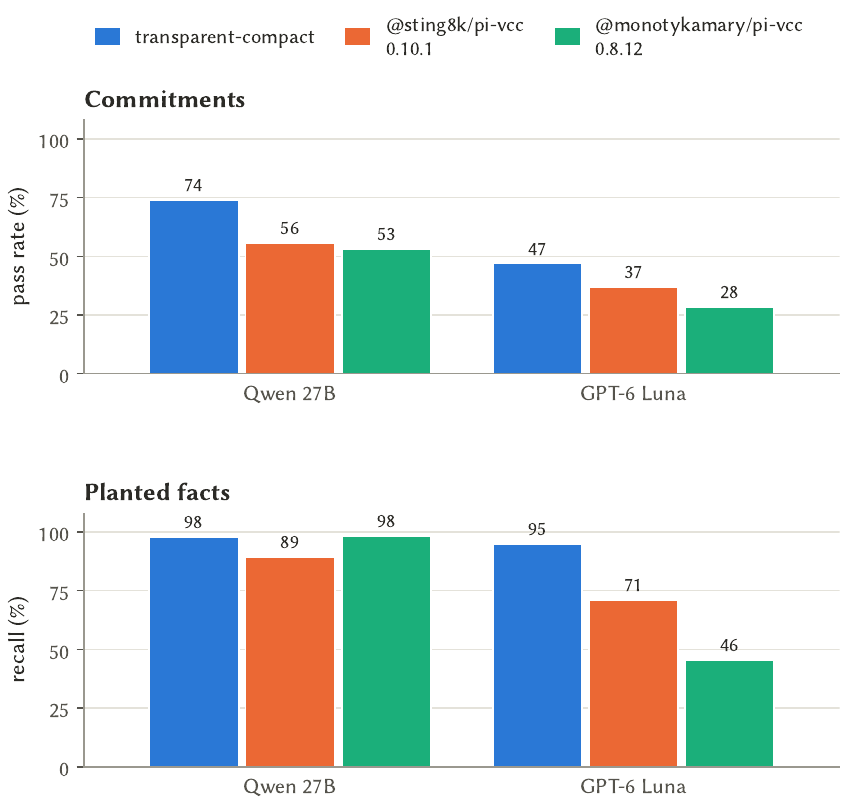}
	\caption{\textbf{Exploratory results comparing the two main versions of pi-vcc} Same parameters as in Figure \ref{fig:threeway}.}
	\label{fig:vcc appendix}
\end{figure}

Preliminary testing indicated that @sting8k's more complex retrieval often caused agents to make excessive retrieval calls while looking for specific information in our benchmarks -- often hitting a 30 minute timeout for individual probes, especially in the planted facts study. We determined that the relative simplicity of @monotykamary's retrieval made it a more suitable comparison for isolating different compaction principles of \vcc{} vs \vb{}. 
In practical usage however, @sting8k is the more mature and popular version, and it appears likely that is the more performant version. Figure \ref{fig:vcc appendix} shows results of exploratory analysis indicating that performance in the benchmark does differ between the two versions, especially with the Luna model, but the \vb{} method still maintains an edge in these studies.

\end{document}